\documentclass{article}
\usepackage{spconf,amsmath,amssymb,graphicx,multirow}

\title{Anatomical labeling of brain CT scan anomalies using multi-context nearest neighbor relation networks}
\name{Srikrishna Varadarajan, Muktabh Mayank Srivastava, Monika Grewal\textsuperscript{*}, Pulkit Kumar\sthanks{Authors contributed equally}}
\address{ParallelDots, Inc.\sthanks{www.paralleldots.xyz}\\
{\tt\small{\{srikrishna, muktabh, monika, pulkit\}@paralleldots.com }}}
\begin{document}
%
\maketitle
\begin{abstract}
This work is an endeavor to develop a deep learning methodology for automated anatomical labeling of a given region of interest (ROI) in brain computed tomography (CT) scans. We combine both local and global context to obtain a representation of the ROI. We then use Relation Networks (RNs) to predict the corresponding anatomy of the ROI based on its relationship score for each class. Further, we propose a novel strategy employing nearest neighbors approach for training RNs. We train RNs to learn the relationship of the target ROI with the joint representation of its nearest neighbors in each class instead of all data-points in each class.  The proposed strategy leads to better training of RNs along with increased performance as compared to training baseline RN network.
\end{abstract}
\section{Introduction}
\label{sec:intro}

Computed Tomography (CT) is primarily used for detection and diagnosis of many trauma related abnormalities in brain e.g. hemorrhage, infarct, edema, and skull fracture. Anatomical localization of the underlying pathologies is often required to gain further insights and plan treatment procedures. The poor anatomical contrast of CT images and non-availability of CT anatomical atlas has so far limited the development of automated methods for anatomical labeling of CT scan images. Traditional image processing based approaches for anatomical labeling of medical images are not only time and computation intensive, but also often rely on availability of multi-modality scans. These limitations render them infeasible for clinical use. The essential characteristics for an automated anatomical labeling solution for CT scan images to be deployable in real-world can be underlined as follows: It should not rely on multi-modality scans, should provide fast inference even with low computational resources that are typical of clinics, and finally, the development cost should not outweigh the delivered benefits. 
Considering above mentioned characteristics, we have sought development of an automated anatomical labeling method of pathological regions in brain CT. First, we have prepared a dataset for supervised learning of anatomical labels for underlying pathologies seen in brain CT using least possible resources. We utilized a dataset that was primarily labeled for supervised learning of multiple pathologies in brain CT scan and anatomical labels were sparsely available as an auxiliary information. Therefore, we obtained the dataset for no additional annotation cost, which otherwise would have been hundreds of hours work by neuroanatomy experts. This approach works on the valid assumption that the practical applications of automated anatomical labeling of CT are limited to localization of pathologies in brain. 
Further, we have implemented a deep learning approach that learns effectively despite small dataset and heavy class imbalance. We propose a cascaded architecture that learns robust embedding of the pathological region of interest (ROI) in first stage, and predicts the corresponding anatomical label in second stage. Our implementation utilizes three approaches to increase performance of model, while dealing with issues of small dataset and class imbalance:-

\begin{enumerate}
	\item We utilize a two-path network at first stage of the architecture, which combines both local and global contexts to learn robust embeddings of the ROI. We refer to this network as BaseNet.
	\item We investigate the use of Relation Networks (RNs) \cite{DBLP:journals/corr/RaposoSBPLB17} to exploit intra-class relations between data points of same class and discriminate them efficiently from data points belonging to different classes.
	\item We propose a novel training strategy employing nearest neighbors approach to increase performance of RNs. We train the RNs such that the network learns more coherent representation of data points belonging to same class and more discriminative representation of data points belonging to different classes.
\end{enumerate}

\begin{figure*}[t]
\centering
{\includegraphics[width=0.7\textwidth]{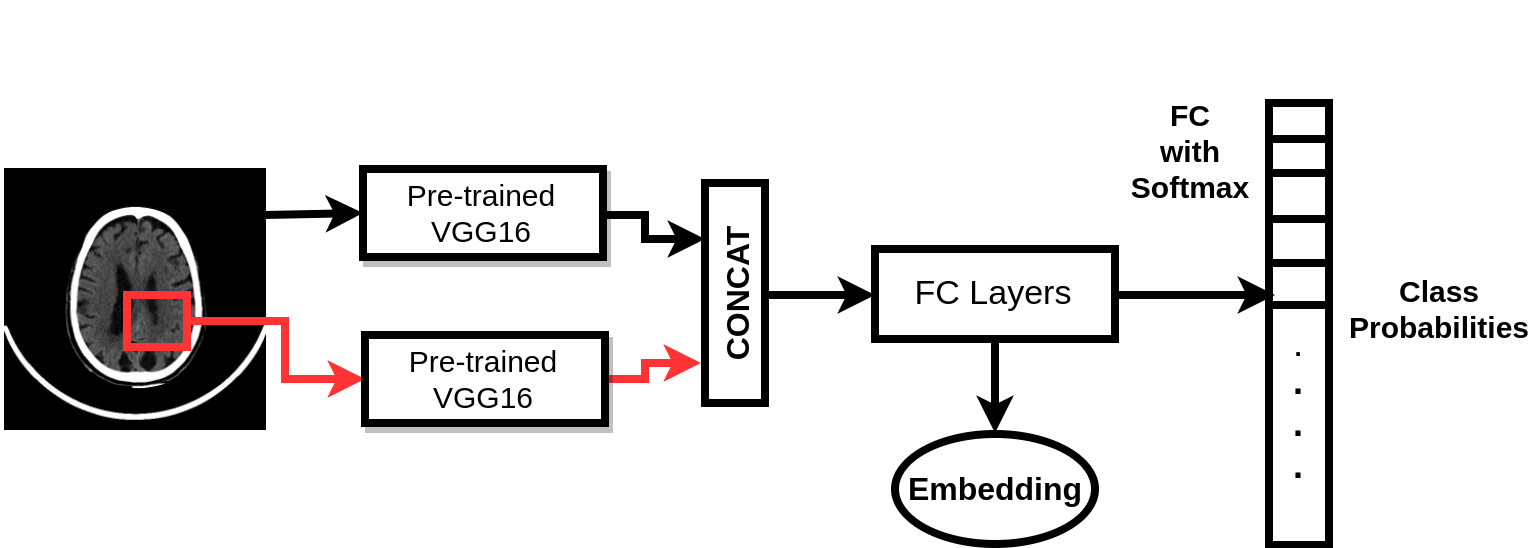}}
\caption{BaseNet}
\label{Fig:basenet}
\centering
\end{figure*}

\section{Method}
\label{sec:method}

The key component of our approach is the way Relation Networks are trained for multi-class classification problem with dataset comprising of class imbalance. RNs are typically implemented as two cascaded networks of fully connected (fc) layers: $g(\theta)$ and $f(\phi)$, wherein, $g(\theta)$ learns to compute joint representation of a  reference set of objects by exploiting relations between them and $f(\phi)$ learns to predict whether an input object representation belongs to the reference set of objects. However, the broad definition of RNs makes them inherently suitable for all kind of multi-class classification problems, where first network of fc layers computes the embedding of a reference class and the second network predicts whether an input object belongs to the same class or not. 

In a multi-class classification problem with heavy class imbalance, training RNs for each data-point (all-to-all version as mentioned in \cite{DBLP:journals/corr/SantoroRBMPBL17}) leads $g(\theta)$ to be biased towards relations in the majority class. Further, the training may be overwhelmed by non-related examples from majority classes. To alleviate this, we employ a novel training strategy utilizing nearest neighbors approach. Let us consider an embedding corresponding to a target data point, and a set of embeddings corresponding to a reference class to be inputs to the RNs. Instead to sending all the reference class embeddings to $g(\theta)$, we instead extract a fixed set of nearest neighbors of the target embedding from the reference class embeddings and send it as input to $g(\theta)$. This leads $g(\theta)$ to return a joint embedding of the reference class (henceforth, referred to as reference embedding) from data points, which are related to target embedding. Consequently, the inputs to $f(\phi)$ would be the target embedding and the reference embedding with high similarity metric. For a case where both the target and the reference embedding belong to the same class, $f(\phi)$ would be trained on a high threshold of similarity between two; whereas for a case of the target and reference embedding belonging to different classes, $f(\phi)$ would be forced to learn more discriminative features separating the two classes.

The proposed strategy can also be interpreted as a way of hard example mining for different classes and easy example mining for same class. The underlying assumption is that the representations of each class would be very different from each other. However, there might still be a lot of examples lying on the cluster boundaries; these examples would be nearest neighbors to the examples from neighboring class. We refer to these samples as hard negatives as it is more difficult for the network to classify them as belonging to different class as compared to those farther away from boundaries. While training for different classes, we focus only on these examples. Similarly for within class samples, there may be many examples lying on the cluster boundary, which may be attributed to either label noise or ambiguous samples. Training positive class on these samples may contribute to learning incoherent class representation. Training with nearest neighbors approach ensures that the representation of a single class is more coherent. In essence, the proposed strategy helps the network learn more discriminative boundaries between classes while avoiding overfitting to individual class features.

\begin{figure*}[t]
\centering
{\includegraphics[width=0.93\textwidth]{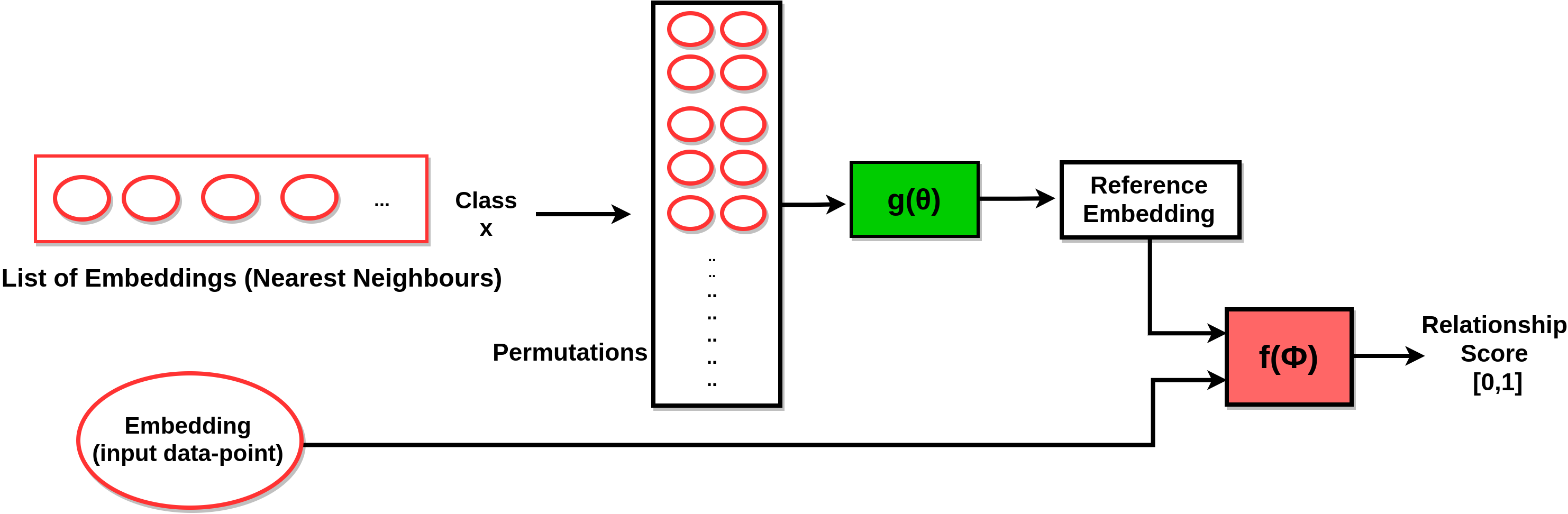}}
\caption{Implementation of RL-Nearest neighbor approach}
\label{Fig:rlnn}
\centering
\end{figure*}

\subsection{RELATED WORK}
\label{ssec:relatedwork}

Automated anatomical labeling methods based on image processing and deep learning have been explored for MRI \cite{Brebisson_2015_CVPR_Workshops,roy2017error,Wang2013Multi-atlasFusion} and PET imaging. Most of these methods are based on segmentation of 3D brain image into different anatomical regions. However, there does not exist any approach addressing anatomical labeling in CT scan images as such. 

Relation Networks have been proposed and used for relational reasoning  \cite{DBLP:journals/corr/RaposoSBPLB17,DBLP:journals/corr/SantoroRBMPBL17}, where the deep learning model is required to extract relations between different objects for prediction. However, recently  \cite{DBLP:journals/corr/abs-1708-06819} have used RNs for few shot learning of multi-class classification task. Our approach of using RNs is, therefore, more similar to that proposed by \cite{DBLP:journals/corr/abs-1708-06819}. 

The idea of combining local and global context is quite intuitive and  logical. There exists substantial amount of prior work that uses this concept in various tasks \cite{Murphy2006,Wright2016IncorporatingSlides,Zhao2015SaliencyLearning,Havaei2017BrainNetworks}. One very relevant work is by \cite{Havaei2017BrainNetworks}, where they do dense segmentation of each pixel to localize the brain tumours by using a local and global context around each pixel. Further, nearest neighbor is a famously used machine learning technique that is used for a plethora of unsupervised as well as supervised tasks. However, to the best of our knowledge, this is the first work utilizing nearest neighbor approach as a method for example mining (hard as well as easy) while training RNs.

\section{Implementation}
\label{sec:implementation}

\subsection{Dataset}
\label{ssec:dataset}

The dataset comprised of 15,425 2D slices corresponding to 216 CT scans acquired in two local hospitals. The dataset was obtained after the approval of ethics committees at hospital level. The data had pathological regions annotated along with underlying anatomy. A total of 15 anatomical labels were present in entire dataset. The number of 2D slices present for each of the 15 anatomical labels present in the dataset are mentioned in Table \ref{Table:dataset}. The dataset was separated into train and test set at CT level with 80:20 ratio, which corresponded to 173 CTs for training and 43 CTs for testing.

\begin{table}[!h]
\begin{tabular}{|l|l|l|l|}
\hline
\textbf{SNo} & \textbf{Anatomy Label}  & \textbf{\# of slices} & \textbf{Percentage} \\ \hline
1   & along falx/tentorium   & 1025              & 6.65       \\ \hline
2   & basal cisterns         & 504               & 3.27       \\ \hline
3   & brainstem              & 95                & 0.62       \\ \hline
4   & cerebellum             & 236               & 1.53       \\ \hline
5   & ethmoidal              & 113               & 0.74       \\ \hline
6   & frontal region         & 4263              & 27.64      \\ \hline
7   & gangliocapsular region & 146               & 0.95       \\ \hline
8   & maxillary              & 622               & 4.04       \\ \hline
9   & occipital region       & 760               & 4.93       \\ \hline
10  & parietal region        & 2341              & 15.18      \\ \hline
11  & sphenoid               & 254               & 1.65       \\ \hline
12  & sulcal spaces          & 1026              & 6.66       \\ \hline
13  & temporal region        & 2520              & 16.34      \\ \hline
14  & thalamus               & 51                & 0.34       \\ \hline
15  & ventricular system     & 1469              & 9.53       \\ \hline
    & Total Slices           & \textbf{15425}             &            \\ \hline
\end{tabular}
\caption{Dataset information}
\label{Table:dataset}
\end{table}

\subsection{Preprocessing} 
\label{ssec:preprocessing}
Preprocessing included resampling to achieve isotropic resolution and fixed field-of-view in-plane, thresholding Hounsfield Units (HU) to brain window, and converting all pixel intensities to the range 0 to 1. The intensity values in the regions corresponding to pathologies were replaced by uniform gray scale values so as to remove any pathology texture related information from the images.

\begin{table*}[t]
\resizebox{\textwidth}{!}
{\begin{tabular}{|l|l|l|l|l|l|l|l|l|l|l|l|l|l|}
\hline
                           &                    & 1     & 2     & 5     & 6     & 8     & 9     & 10    & 11    & 12    & 13    & 15    & Mean \\ \hline
\multirow{3}{*}{Precision} & BaseNet            & 0     & 0     & 0     & 0.961 & 0     & 0     & 0.401 & 0     & 0     & 0.246 & 0.367 & 0.422   \\ \cline{2-14} 
                           & BaseNet+RN       & 0.609 & 0.128 & 0     & 0.762 & 0.772 & 0.477 & 0.466 & 0.814 & 0     & 0.668 & 0.528 & 0.567   \\ \cline{2-14} 
                           & BaseNet+RN+NN & 0.552 & 0.121 & 1     & 0.82  & 0.773 & 0.597 & 0.446 & 0.821 & 0.252 & 0.657 & 0.411 & 0.602   \\ \hline
\multirow{3}{*}{Recall}    & BaseNet            & 0     & 0     & 0     & 0.232 & 0     & 0     & 0.452 & 0     & 0     & 0.976 & 0.359 & 0.332   \\ \cline{2-14} 
                           & BaseNet+RN       & 0.508 & 0.121 & 0     & 0.885 & 0.992 & 0.205 & 0.601 & 0.846 & 0     & 0.777 & 0.55  & 0.635   \\ \cline{2-14} 
                           & BaseNet+RN+NN & 0.519 & 0.097 & 0.033 & 0.817 & 1     & 0.371 & 0.59  & 0.884 & 0.091 & 0.694 & 0.539 & 0.615   \\ \hline
\multirow{3}{*}{F1-score}  & BaseNet            & 0     & 0     & 0     & 0.374 & 0     & 0     & 0.425 & 0     & 0     & 0.393 & 0.363 & 0.372   \\ \cline{2-14} 
                           & BaseNet+RN       & 0.554 & 0.124 & 0     & 0.819 & 0.868 & 0.287 & 0.525 & 0.83  & 0     & 0.718 & 0.539 & 0.594   \\ \cline{2-14} 
                           & BaseNet+RN+NN & 0.535 & 0.108 & 0.064 & 0.818 & 0.872 & 0.458 & 0.508 & 0.851 & 0.134 & 0.675 & 0.466 & 0.597   \\ \hline
\end{tabular}}
\caption{Performance comparison between BaseNet, BaseNet+RN, BaseNet+RN+NN}
\label{Table:results}
\end{table*}

\subsection{Architecture}
\label{ssec:architecture}
The architecture consists of a baseline two-path network cascaded with Relation Networks (Fig \ref{Fig:rlnn}). The baseline network, referred to as BaseNet (Fig \ref{Fig:basenet}), extracts features corresponding to local and global context along its two processing paths. The feature vector for both the contexts are obtained from the output of pool5 layer of pre-trained\footnote{https://github.com/pytorch/vision\#models} VGG16 \cite{DBLP:journals/corr/SimonyanZ14a}. The obtained local and global features are later concatenated and passed through a network of 3 fc layers which are trained from scratch for multi-class classification. The later part of the cascaded network consists of RNs composed of two networks of 3 and 2 fc layers representing $g(\theta)$ and $f(\phi)$ implementations, respectively. BaseNet and RNs are trained separately, with the output of BaseNet's second last fc layer passed as input embedding to the RNs. Dropout of 0.5 was used between fc layers to avoid overfitting. The implementation was done in PyTorch.

\subsection{Training}
\label{ssec:training}
Since the class distribution in the dataset is skewed, the representations learnt by training BaseNet on whole dataset would have been biased towards majority classes. In order to avoid this, we over sampled the minority classes by data augmentation and undersampled the majority classes during BaseNet training. We used stochastic gradient descent optimization with a learning rate of 1e-3, momentum of 0.9, and weight decay of 5e-4 for training. BaseNet and RNs were respectively trained for 30 and 15 epochs, which took 33 minutes and 67 hours, respectively. This is to be noted that although RN contains less number of parameters, training time for RN is still high. This is because of the increase in effective data points while training as we take the relationship of each datapoint with all the classes.

\subsection{Testing}
\label{ssec:Testing}
During testing, the nearest neighbors for target embedding were fetched from the training dataset. The relationship of the embedding with each class was determined as a relationship score lying in the range of 0 to 1. The class which the embedding belonged to, was determined by taking the argmax of all the relationship scores. 

We calculated precision, recall, and F1-score for each class as evaluation metrics using sklearn \cite{scikit-learn} package. The mean metrics were calculated by weighing individual class metrics by number of true instances in the class. This method accounts for underlying class imbalance while reporting average metrics.  

\section{Results}
\label{sec:results}

We achieve 0.602, 0.615, and 0.597 as mean precision, recall, and F1-score as shown in Table \ref{Table:results}. We haven't shown the evaluation metrics on 4 classes as they contained very small number of samples in the test set (as few as 5 slices). However, the mean metrics shown is the average are over all the classes. None of the classes were removed due to its low sample size. 

The results clearly demonstrate the gain in performance by the use of RNs. Further, RN + Nearest neighbor (NN) approach shows marginal increase in F1-score and precision. This should be noted that although the performance gain achieved by RN + NN approach seems to show only marginal gain, in practice, this amounts to better performance in minority classes. The reason behind the increased performance in these classes not being reflected in mean score is because the total representation of the said classes in the complete test set is very small. Therefore, it can be said that training RNs with NN approach leads to more balanced performance across all the classes and thus reducing the effect of class imbalance.

To gain further insights about the relative performance of BaseNet + RN and BaseNet + RN + NN models, we compared the mean F1-scores of both the models under equal training time conditions. At the end of 5 epochs, the mean F1-score of BaseNet + RN was 0.53, whereas for BaseNet + RN + NN was 0.59. This shows that BaseNet + RN + NN trains much faster than BaseNet + RN model, while also learning on minority classes.

\section{Conclusion}
\label{conclusion}
We have proposed a deep learning method for automated anatomical labeling of an ROI in brain CT scans. We have employed a cascaded architecture, which computes a representation of the ROI by combining local and global context in first stage, and uses Relation Networks to predict the underlying anatomy from the representation in second stage. Further, we have proposed a novel training procedure utilizing nearest neighbor approach to train RNs and argue that it employs a kind of hard as well as easy example mining to help in robust training of the network. We also reported the test results for all the approaches used by us and compared their relative performance. However, we would want to mention that the presented approach still does not perform well on some minority classes, for which the number of samples were very small. We believe that further increase in the number of samples would help us achieve better performance on the minority classes in future. 

In summary, we have provided a low resource yet efficient solution for anatomical labeling of a given ROI in CT scan images. However, the methods reported in this paper are not limited to the said application area and as such can be applied to various other fields facing similar issues. 


\bibliographystyle{IEEEbib}
\bibliography{Mendeley,refs}

\end{document}